\documentclass{article}
\newcommand{\final}{1}
\usepackage[final]{neurips_2019}

\usepackage{natbib}

\usepackage{url}
\usepackage{color}
\usepackage{parskip}
\usepackage{ifthen}
\usepackage{float}
\usepackage{alltt}
\usepackage{amsmath}
\usepackage{amssymb}
\usepackage{amsthm}

\theoremstyle{definition}
\newtheorem{definition}{Definition}[]

\theoremstyle{remark}
\newtheorem*{remark}{Remark}

\usepackage{color}
\usepackage{parskip}
\usepackage{ifthen}
\usepackage{float}
\usepackage{alltt}
\usepackage{amsmath}
\usepackage{bbm}
\usepackage{newlfont} 
\usepackage{floatflt}
\usepackage{fixltx2e}
\usepackage{subfig} 
\usepackage{multirow}
\usepackage{booktabs}
\usepackage{cleveref}
\usepackage[export]{adjustbox}
\usepackage{mathtools, cuted}
\usepackage{CJKutf8} 

\usepackage{float}
\restylefloat{table}

\usepackage{algorithm}
\usepackage{algpseudocode}
\makeatletter
\def\BState{\State\hskip-\ALG@thistlm}
\makeatother

\def\textrightarrow{$\rightarrow$}

\newcommand{\Caption}[2]{\caption[#1]{{\em #1} #2}}

\definecolor{SithColor}{rgb}{0.7,0,0} 
\newcommand{\martin}[1]{{\color{SithColor} Martin: #1 $\qed$}}
\definecolor{ConsularColor}{rgb}{0,0.4,0} 
\definecolor{GuardianColor}{rgb}{0,0,0.8} 
\newcommand{\jialin}[1]{{\color{GuardianColor} Jialin: #1 $\qed$}}
\newcommand{\junpei}[1]{{\color{ConsularColor} Junpei: #1 $\qed$}}
\newcommand{\warning}[1]{{\it\color{red} #1}}
\newcommand{\note}[1]{{\it\color{blue} #1}}
\newcommand{\nothing}[1]{}

\definecolor{AudioColor}{rgb}{0.56,0.34,0.62}

\definecolor{figred}{rgb}{1,0,0}
\definecolor{figgreen}{rgb}{0,0.6,0}
\definecolor{figblue}{rgb}{0,0,1}
\definecolor{figpink}{rgb}{1,0.63,0.63}

\ifthenelse{\equal{\final}{1}}
{
\renewcommand{\martin}[1]{}
\renewcommand{\jialin}[1]{}
\renewcommand{\junpei}[1]{}
\renewcommand{\warning}[1]{}
\renewcommand{\note}[1]{}
}
{
\renewcommand{\note}[1]{}
}

\newcommand{\pseudocode}{Algorithm}
\floatstyle{plain}
\newfloat{algorithm}{tbhp}{lop}
\floatname{algorithm}{\pseudocode}

\newcommand{\filename}[1]{\url{#1}}
\newcommand{\foldername}[1]{\url{#1}}

\let\oldparagraph\paragraph

\renewcommand{\paragraph}[1]{\oldparagraph{\textbf{#1}.}} 

\ifdefined\email
\else
\newcommand{\email}[1]{\url{#1}}
\fi

\newcommand{\AlgoName}{\textbf{ADS} }

\graphicspath{
{figs/handdrawn/}
{figs/raster/}
{../../Figures/}
}

\begin{document}

\title{An Active Approach for Model Interpretation}

\author{Jialin Lu \qquad
Martin Ester \\
School of Computing Science\\
Simon Fraser University
}





\maketitle

\section{Introduction}
\note{

\textbf{
What problem we are trying to solve.
Why it is important, and why people should care.
}

Writing research documents, such as conference/journal/white papers, patent applications, grant proposals, and course reports, is a core activity for many poor souls including professors, researchers, engineers, and students.
Since we must do it one way or another, sooner or later, we might as well do it as happily and effectively as possible.
Here are my personal practices and experiences, which I have repeatedly shared with my collaborators and students throughout all these years  to the point that I finally decided to write up to save the troubles.

This paper:
Introduce and describe functional parcellation, mention a wide list of applications and then emphasize on brain parcellation. Describe how it is important.
}
Model interpretation, or explanation of a machine learning classifier, aims to extract generalizable knowledge from a trained classifier into a human-understandable format, for various purposes such as model assessment, debugging and trust. From a computational viewpoint, it is formulated as approximating the target classifier using a simpler interpretable model, such as rule models like a decision set/list/tree.
Often, this approximation is handled as standard supervised learning and the only difference is that the labels are provided by the target classifier instead of ground truth. This paradigm is particularly popular because there exists a variety of well-studied supervised algorithms for learning an interpretable classifier.

However, we argue that this paradigm is suboptimal for it does not utilize the unique property of the model interpretation problem, that is, the ability to generate synthetic instances and query the target classifier for their labels. We call this the \emph{active-query} property, suggesting that we should consider model interpretation from an active learning perspective.
Following this insight, we argue that the \emph{active-query} property should be employed when designing a model interpretation algorithm, and that the generation of synthetic instances should be integrated seamlessly with the algorithm that learns the model interpretation. In this paper, we demonstrate that by doing so, it is possible to achieve more faithful interpretation with simpler model complexity.
As a technical contribution, we present an active algorithm \textbf{A}ctive \textbf{D}ecision \textbf{S}et Induction (\textbf{ADS}) to learn a decision set, a set of if-else rules, for model interpretation.
\AlgoName performs a local search over the space of all decision sets.
In every iteration, \AlgoName computes confidence intervals for the value of the objective function of all local actions and utilizes active-query to determine the best one to apply.

\note{
\textbf{
What prior works have done, and why they are not adequate.
(Note: this is just high level big ideas. Details should go to a previous work section.)
}
}
\note{
\textbf{
What our method has to offer, sales pitch for concrete benefits, not technical details.
Imagine we are doing a TV advertisement here.
This document provides a template for writing research papers, and describes how to manage project progress via iterating drafts.
}
Doing this correctly can help you achieve productive research and happy life, e.g. working anywhere anytime without synchronous meetings with your collaborators.
}
\note{
\textbf{
Our main idea, giving people a take home message and (if possible) see how clever we are.
}
I learned from my PhD adviser to start writing from the moment I have the faintest idea, and gradually update the drafts to reflect progress.
During my first job with NVIDIA, I learned why and how revision control is essential for sharing and editing source files, especially among large engineering teams.
The iterative nature of writing and coding matches well with the capability of revision control.
Thus I prefer Latex + git/svn (e.g. github, bitbucket, and the svn server under my hosting service), even though I can still manage with MS Word + cloud drive (e.g. Google drive, Box, and Dropbox).
A good computer scientist writes papers like programs, and manages Latex files like source codes.
These repos are external memories and communication mediums for the collective brains of my teams.
}
\note{
Our algorithms and methods to show technical contributions and that our solutions are not trivial.
\begin{figure}[htb]
  \centering
  \subfloat[raster]{
    \label{fig:example:raster}
    \includegraphics[width=0.4\linewidth]{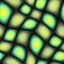}
  }
  \subfloat[vector]{
    \label{fig:example:svg}
    \includegraphics[width=0.58\linewidth]{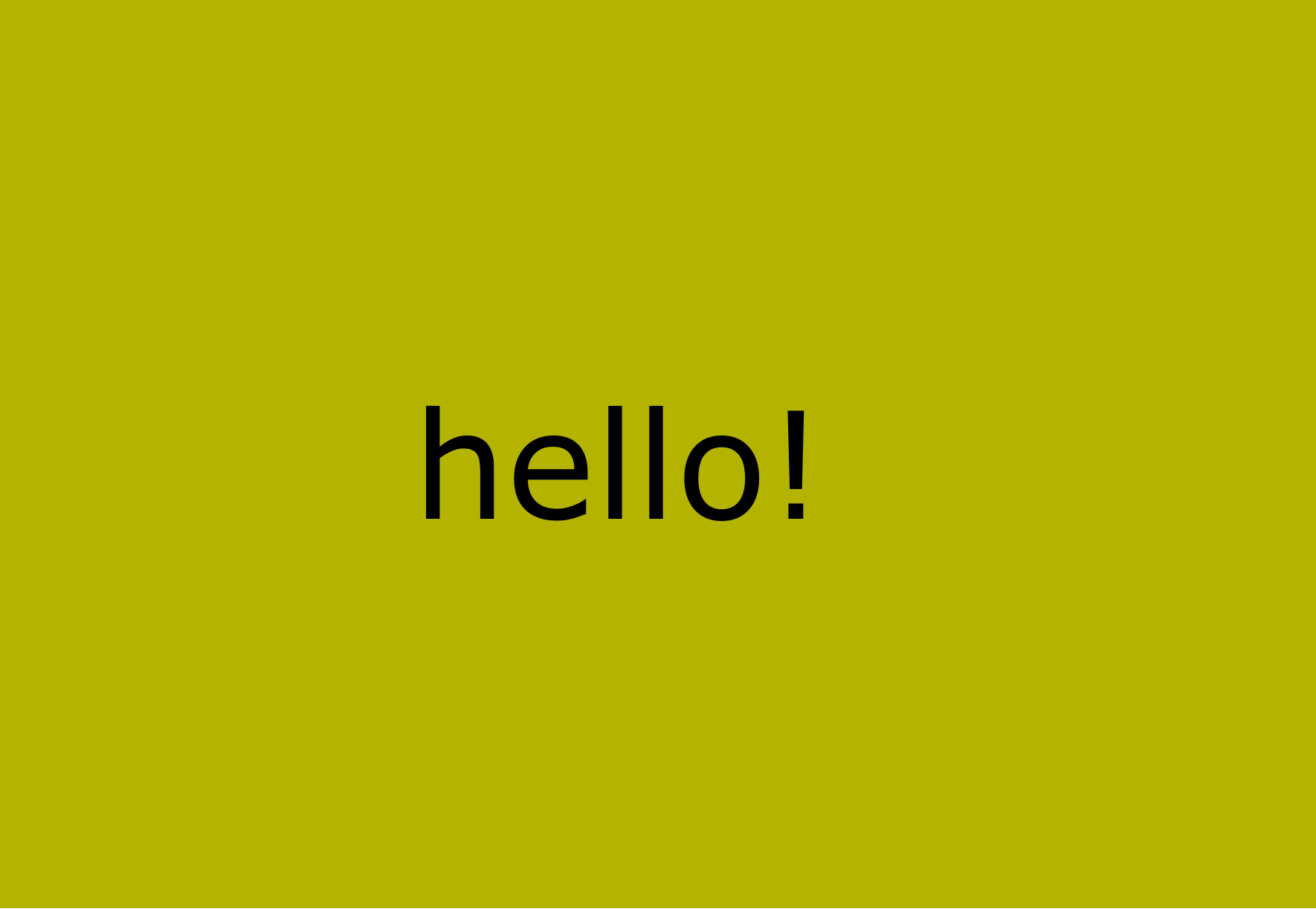}
  }

  \subfloat[vector + raster]{
    \label{fig:example:svg_img}
    \includegraphics[width=0.9\linewidth]{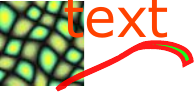}
  }
 \Caption{Example figure.}
 {%
\subref{fig:example:raster} is a raster image and \subref{fig:example:svg} is a vector graphics.
Never, ever, rasterize vector graphics unless you want large size and low quality files.
We can also combine vector and raster graphics as in \subref{fig:example:svg_img}.
Use Inkscape to import  an image into an svg file, and draw over it using whatever stuff like texts or strokes.
Make sure the image is linked not embedded to avoid duplicating the image, and the link should be relative in the same path \cite{StackExchange:2011:HLI}.
 }
 \label{fig:example}
\end{figure}

Translating thoughts into words, diagrams, equations, pseudo-codes, or codes is a good way to clarify and consolidate.
If you are writing a graphics/HCI paper, design the figures so that they alone can provide a high level picture of the main points.
As exemplified in \Cref{fig:example}, a figure can contain images, drawings, or combinations.

}
\note{
\textbf{
Results, applications, and extra benefits.
}
You can use the associated files as a starting template for your research papers.
Start with the \filename{Makefile}, and notice the two build targets: final for official submission and public disclosure, and draft for internal sharing among your collaborators (including yourself).
}

\section{Related work}
\label{sec:prior}
\note{
Cite all relevant works, and how yours stands on the shoulder of these previous giants.
Use a bibtex file such as \filename{paper.bib} and you can cite relevant works such as \cite{Sun:2016:MVP}.
}
We discuss global model interpretation methods that approximate the global behavior of a target classifier and we further restrict ourselves to rule-based methods.

First introduced is the mainstream of model interpretation methods which we refer to as \emph{passive} methods, in the sense that it approximates a fixed-sized dataset labelled by the target classifier, rather than the target classifier itself.
\citet{Jure:2017:BETA,lakkaraju:2019:faithful} learns a decision set for model interpretation by jointly optimizing interpretability and approximation faithfulness on the relabelled dataset.
Following this widely-cited approach, a variety of alternative algorithms for learning an interpretable classifier \citep{Lakkaraju:2016:IDS,Wang:2017:BRS,Yang:2017:SBRL} can be applied in the context of model interpretation.
But the problem is that the given dataset, in practice, may not be good enough to reveal the true decision boundary of the target classifier and
and thus whether this interpretable simpler model can faithfully approximate the target classifier is in question.
The approaches of \citet{Bertini:2018:RuleMatrix,Hamfelt:2019:RICE,Sushil:2018:RuleInduction} first augment the dataset and then apply the main algorithm for learning an interpretable model. For example, \citet{Bertini:2018:RuleMatrix} first estimate the data distribution and then sample an arbitrarily large dataset.
\emph{Passive} methods can utilize new data instances, but only as preprocessing which is separate from the main learning algorithm, and thus we argue is suboptimal.

Another line of work, which we call \emph{bottom-up} methods, applies two-phase local-to-global algorithms \citep{Dino:2018:local_first} leveraging the relatively well-studied local interpretation methods \citep{Ribeiro_2016,Ribeiro:2018:Anchor,LORE}. The idea is to first construct a local interpretation for each instance in the dataset, and then select a subset of them to produce a global interpretation.

Lastly, we introduce the work closest to ours, DT-Extract \citep{Bastani:2017:active_tree}, as it integrates active-query seamlessly into the algorithm. It modifies the classic decision tree algorithm CART by generating new instances at each node to select the best split and thus better approximates the target classifier.
We believe this tree-growing approach has been re-invented many times \citep{Bastani:2017:active_tree,Craven:1996:TREPAN,Breiman:1996:born_again} though in different contexts
and suffers from the inherent drawback of decision tree methods, that its approximation does not consider interpretability explicitly.
In order to be interpretable, the number of node splits is hard-coded and thus the number of leaf nodes is controlled. We refer to these methods as \emph{top-down} methods.

\section{Definitions and Problem
Formulation}
\label{sec:setting}

\note{

People like to know what you have done before how it is done.

For HCI papers it is common to start with the user interface or usage scenarios.
Even for a very theoretical paper, we can summarize the main equations/methods/applications/implications here, before diving into technical details.

}

We are given a pre-trained binary classifier $f$ that receives an instance $x$
and returns a prediction $f(x) \in \{0,1\}$. Let $X = \{x\}^n =  \{ (a_1,a_2,...,a_m)\}^n$
represent an instance set of $n$ instances with $m$ attributes.
The input space is defined as $D_x = \{dom(a_i)\}^m $.
The form of model interpretation is a decision set, consisting of a set of if-then rules.
It predicts the positive class if at least one of the rules is satisfied and predicts the negative class otherwise.
    A condition (clause) $c$ consists of an attribute and a range of values this attribute can take.
    Conditions on continuous attributes are specified as bounded intervals, for
    example, $\text{`price'} \in [2.33, 10]$, and conditions on categorical features as a list
    of values, for example, $\text{`state'} \in \{\textit{California},\textit{Texas}\}$.
A rule is a conjunction of conditions
$r =
c_1 \land c_2 \land ... \land c_{|r|} $.
We use $r(\cdot)$ to represent the Boolean function  $r(\cdot): D_x \mapsto \{0,1\}$, which computes whether or not an instance $x$ satisfies $r$.
\begin{definition}
A decision set is a set of rules, denoted
as $S = \{r_1,r_2,...\}$.
An instance is predicted as positive if it
satisfies at least one of the rules, as defined by
$S(\cdot)$:
$S(x) =
\begin{cases}
  1 & \exists r \in S, r(x) = 1 \\
  0 & \text{otherwise.}
\end{cases}
$
\end{definition}

Our goal is to determine a decision set $S$ that maximizes a simple objective function
$$Q(S)  = \theta_S - \lambda |S| $$
where the first term $\theta_S = E_{x \sim P(X) }[ \mathop{\mathbbm{1}}_{f(x) = S(x)}   ] $ measures the faithfulness of the approximation of our
decision set to the target classifier, defined as the expected accuracy over the input space. $|S|$ denotes the
number of rules in $S$ as a measure of
interpretability. $\lambda$ is a tunable hyper-parameter set
to $0.01$ by default, meaning that a $1\%$
accuracy improvement is worth adding an extra rule.
\begin{remark}
The design of the objective function is not the focus of this paper.
Any reasonable objective could be used instead, such as \cite{Lakkaraju:2016:IDS,lakkaraju:2019:faithful,Wang:2017:BRS}, as interpretability comes in different forms in different domains \citep{Freitas_2014,Huysmans_2011}.
\end{remark}

\section{Active Decision Set Induction for model interpretation}
\label{sec:method}
\note{
Written words are more concrete than ideas in your head.
Math equations and pseudo-codes are even more concrete than words, and can help, even though not essential, for exposition.
Programs that produce the actual results are the most concrete, but the amount and level of information is not suitable for a research paper.
I like to write down math and pseudo-code first as a way to design the algorithms before actual implementation.
}
\note{
\subsection{Code}
\label{sec:code}

Pseudo-code can help summarize and explain the algorithms, but only if presented with sufficient clarity and simplicity.
}
Given a target classifier $f$ and a data set of real data instances $X$, our goal is to find a decision set that maximizes the objective function $Q$.
We propose a local search algorithm called Active Decision Set Induction (\textbf{ADS}) which utilizes active query to determine the best action in each iteration.
The decision set $S$ is initialized as empty. In each iteration, a set of actions $A$ that locally modify the current decision set is generated, among which the best action $a \in A$ is determined and then applied to maximize $Q$.
The actions considered at each iteration are modifications of the current decision set
of the following types, adding/removing a rule, adding/removing/modifying a condition of a rule.
After actions are generated, active query is utilized to determine the increase of objective function after applying an action.
To escape from local optima, a simple $\epsilon$-greedy strategy is employed: with probability $\epsilon$, the algorithm chooses a random action.

\begin{algorithm}
\caption{The \AlgoName algorithm}
\label{alg:overall}
\begin{flushleft}
        \textbf{INPUT:} A tuple $(Q,f,X)$  . $Q$ - the objective function, $f$ - the target classifier model, $X$ - the given set of instances\\
        \textbf{Algorithm Parameters}: $\beta$ - the confidence parameter, $\epsilon$ - the randomness for $\epsilon$-greedy strategy, $N_{max}$ - the maximum number of iteration for the local search \\
        \textbf{OUTPUT:} $S$ - the decision set \\

\end{flushleft}
\begin{algorithmic}[1]
\Procedure{Main}{}
\State $S_0 \gets  \emptyset $ \Comment{initialize an empty decision set}
\State $ Y \gets f(X) $ \Comment{query existed data instances}
\State $ X' \gets \emptyset$ , $Y' \gets \emptyset$ \Comment{an empty set of synthetic instances}
\For{ $t =1,2,\dots, N_{max} $ }
\State $A \gets GenerateActions(S_{t-1})$ \Comment{Add/remove a rule, etc}
\State $a^* \gets \mathop{\mathrm{argmax}}\limits_{a \in A} \hat{Q}(a(S_{t-1}) )  $ \label{alg:main:l1}
\State $a' \gets \mathop{\mathrm{argmax}}\limits_{a \in A, a \neq a^* } U_a $ \label{alg:main:l2}
\While { $ L_{a^*} < U_{a'} $  }
    \State $ x_1 \gets GenerateSyntheticInstances(r_{a^*},X)$; $ y_1 \gets f(x_1)$
    \State $ x_2 \gets GenerateSyntheticInstances(r_{a'},X)$; $ y_2 \gets f(x_2)$ \Comment{synthetic instances}

    \State $ X' \gets X' \bigcup \{x_1,x_2\}  $ , $Y' \gets Y' \bigcup \{y_1,y_2\}  $
    \State update $\hat{Q}((S_{t-1}))$ and $\hat{Q}(a'(S_{t-1}))$
    \State $a^* \gets \mathop{\mathrm{argmax}}\limits_{a \in A} \hat{Q}(a(S_{t-1}) )  $
    \State $a' \gets \mathop{\mathrm{argmax}}\limits_{a \in A, a \neq a^* } U_a $
    \Comment{reset $a^*$ and $a'$}
\EndWhile

\State $S_{max} \gets \max ( \hat{Q}(a^*),\hat{Q}(S_{max}) )$
\State $S_t =
\begin{cases}
  a_{random}(S_{t-1}) & \text{with probability } \epsilon \\
  a^*(S_{t-1}) & \text{otherwise.}
\end{cases}
$

\EndFor

\State \Return $S_{max}$
\EndProcedure

\end{algorithmic}
\end{algorithm}




\textbf{When to utilize active query?}
In the framework of local search,  the best action for maximizing the objective is chosen and applied.
For the problem of model interpretation, it is non-trivial to determine the best action,
because the faithfulness term $\theta_S = E_{  x \sim P(x) } [  \mathbbm{1}_{f(x) = a(S)(x)}  ]$ ( the first term of $Q$ ) can only be estimated as $\hat{\theta}_S = E_{ x \in X \cup X' } [  \mathbbm{1}_{f(x) = a(S)(x)}  ]$, where $X'$ is a set of generated synthetic instances.
It is, at this point of time, for determining the best action that active query is utilized.
We borrow ideas from the best arm identification problem for pure-exploration bandits  \citep{audibert:2010:bmi_mab,kalyanakrishnan:2012:LUCB_PAC,kaufmann:2013:information},
particularly LUCB\citep{kalyanakrishnan:2012:LUCB_PAC}. Note that since the rules of $S$ are descriptive, we can use the rules to generate new instances and query the classifier to better estimate $\hat{\theta}_S$, and thus determine the best action.
For estimating $\hat{Q}$ after applying an action $a$ on the current decision set $S$, a confidence interval is computed to construct a lower and an upper bound for $\hat{Q}$.
For some exploration rate $\beta$, the lower bound and upper bound for an action $a$ are computed as:
$$  L_{a}   = \hat{Q}(a(S)) - \beta  \sqrt{ \frac{ \rho_0  }{ \rho (r_a)   } } \text{ , }  U_{a} = \hat{Q}(a(S)) + \beta \sqrt{  \frac{ \rho_0 }{ \rho (r_a)  }} $$
$\beta$ is a hyper-parameter that controlls the confidence level,
and $\rho(r_a) = \frac{N(r_a)}{V(r_a)}$ defines the relative density: $r_a$ is the rule that is added, removed or modified by action $a$,
$N(r_a)$ denotes the number of instances covered by this rule $N(r_a) = | \{x | r_a(x) = 1\} |$
and $V(r_a)$ denotes the volume of input space covered by $r_a$. $\rho_0$ is the pre-computed density of the given dataset over the entire input space.

Utilizing \emph{active-query} exhaustively for each local action is possible, but the total number of actions can be very large, not to mention that we have to do it for each iteration. Therefore, we use an \emph{adaptive sampling scheme} \citep{kaufmann:2013:information} focusing on only two well-chosen candidate actions:
1) the empirically best action that has the highest empirical estimate $a^* \gets \mathop{\mathrm{argmax}}\limits_{a \in A} \hat{Q}(a(S) ) $  and
2) the most optimistic action that has the highest upper bound $a' \gets \mathop{\mathrm{argmax}}\limits_{a \in A, a \neq a^* } U_{a}$.
New instances are generated and queried to update the estimate and the bounds of the objective function for these two candidate actions. This process continues until the lower bound of $a^*$ is higher than the upper bound of $a'$, i.e., $L_{a^*} > U_{a'}$.
If this termination condition is met, then we are confident that $a^*$ is the best action among all other actions $a \in A$ under the exploration rate $\beta$. $a^*$ is then applied, i.e, $S$ is replaced by $a^*(S)$ and the next iteration is started.
\begin{remark}
    When $\beta = 0$, no synthetic instances will be generated and queried as the termination condition will be immediately met. \AlgoName then reduces to a passive approximation.
\end{remark}

\textbf{How to generate instances for active query?}
The descriptive nature of rules gives us an opportunity to generate new instances to query in order to refine the estimate of the value of the objective function after applying an action.
In order to distinguish $\hat{Q} (a^*(S))$ and $\hat{Q} (a'(S))$ to determine the best action, at each round, one synthetic instance is generated for both $a^*$ and $a'$, and their labels are obtained by querying the classifier $f$.
The generation of new instances is formulated as pool-based sampling: we first generate a pool of candidate instances and then select one to label.
For generating a pool of synthetic instances as candidates for action $a$, note that the alternative actions considered at an iteration have only an edit distance of 1 from $S$, we focus on $r_a$, the rule modified by $a$.
A pool of instances $ X_{pool}= \{ x | r_a(x) = 1  \} $ is randomly generated.

We devise a simple pool-generation scheme called Counterfactual Sampling to generate the random pool $ X_{pool}$. We randomly pick instances from $\{x | x \in X \cup X', r_a(x)=0 \}$ and modify them to satisfy $r_a(x)=1$. Instances are modified to satisfy $r_a$ by replacing the value of the specified attributes by the condition of $r_a$ with uniform-randomly sampled values.
For example, given $r_a = \{ \text{`price'} \in [2.33, 10] \}$ specified on the $\text{`price'}$ attribute and a data instance $x = \{ \text{`state'} = \text{`Texas'}  , \text{`price'} = 1  \}$ that $r_a(x)=0$, the modified instance $x'$ will be $\{ \text{`state'} = \text{`Texas'}  , \text{`price'} = 5  \}$ where $5$ is drawn uniformly from $[2.33, 10]$ and now $r_a(x')=1$.
The rationale for this counterfactual sampling is as follows.
To sample $ X_{pool}$, we first tried uniform randomness but it did not work well. Probably because the generated instances are not from the data distribution $P(X)$.
We also tried to first estimate $P(X)$ and then perform rejection-sampling $\{ x |x \sim P(X), r_a(x) = 1 \}$,
which improved the performance only slightly, perhaps because the quality of estimation of $P(X)$ matters and estimating $P(X)$ is already very hard.
The proposed Counterfactual sampling avoids estimating the true data distribution $P(X)$ and still works well to generate instances \emph{as if} they were from $P(X)$.

For selecting an instance to label, one synthetic instance that maximizes the distance to its nearest neighbor among all existing instances satisfying $r_a$ ($\{ x | r_a(x) = 1, x \in X \cup X' \}$) is selected from the candidate pool.
We then query $f$ to get the label and append it to $X'$ to refine $\hat{Q} (a(S))$.


\begin{remark}
While more sophisticated heuristics for uncertainty-based sampling
exist \citep{MacKay_1992,NIPS:2005:straddle}, we choose to use the largest distance to its nearest neighbor as the sampling criteria, not only because most of the uncertainty measures are essentially based on it, but also we find that it works well and is computationally efficient.
\end{remark}

\section{Evaluation}

\note{
Provide evaluation, analysis, comparisons, demos, and applications.

You can more concretely learn everything I described in this document by apprenticing some research projects with me.
}

We run experiments on the Income Census Prediction (Adult) dataset \citep{Kohavi:1996:Adult}, for it is a widely-used and relatively large tabular dataset (48842 instances).
We  use a 90\%-10\% train-test split and train a 5-layer fully-connected Deep Neural Network on the training set. Model interpretation methods are also applied on the train set and evaluated on the test set.

We analyze the dependency of \AlgoName from different choices of the hyper-parameters $\lambda$ and $\beta$.
We run \AlgoName for different values of $\beta$ and report the curve of F1 score v.s. $|S|$  in \Cref{fig:eval:tuning}. For a given value of $\beta$, we vary $\lambda$ to get different numbers of rules. Note that the exploration rate $\beta$ controls the number of active queries:
a larger $\beta$ will use more active queries while $\beta \gets 0$ reduces \AlgoName to passive mode without generating any synthetic instances.
We find that as $\beta$ gets larger, the curve moves towards the top-left corner, indicating that \AlgoName can achieve better approximation with even fewer rules with the help of active query.

We then compare \AlgoName with several baselines on various metrics for both faithfulness (F1 score, precision and recall) and interpretability (number of rules, number of conditions, etc).
For baselines, we consider one representative method for each of three categories (\emph{passive},\emph{bottom-up}, \emph{top-down}) discussed in Related Work. For passive methods, we choose SBRL+ \citep{Bertini:2018:RuleMatrix} which first enlarges the dataset and then learns a Scalable Bayesian Rule List (SBRL) \citep{Yang:2017:SBRL}. We replace SBRL with Bayesian Rule Set (BRS) \citep{Wang:2017:BRS} for a fair comparison and call this BRS+.
For \emph{bottom-up} methods that merge local interpretations into a global one, we choose SP-anchor \citep{Ribeiro:2018:Anchor}.
As a representative of top-down active methods, DT-Extract \citep{Bastani:2017:active_tree} is also compared.
All root-to-leaf paths are extracted as decision rules from the tree. For SP-anchor and DT-Extract, since the number of rules can be controlled through hyper-parameters, we force them to produce the same number of rules as \textbf{ADS}.
The result presented in \Cref{table:eval:overall} shows that \AlgoName significantly outperforms the baselines in all faithfulness metrics (except precision) and in all interpretability metrics.
Compared to \AlgoName non active, \AlgoName greatly improves the recall and F1-score.
SP-anchor and DT-Extract produce rules of very high precision but surprisingly low recall, in some way failing the purpose of approximating the global behavior of the target classifier.


\begin{minipage}{\textwidth}
    \begin{minipage}[b]{0.24\textwidth}
        \centering
        \includegraphics[width=\linewidth]{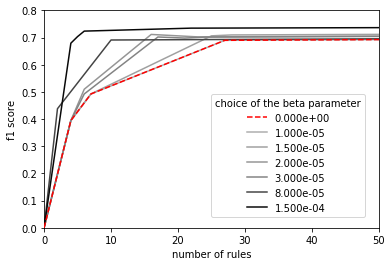}
        \captionof{figure}{Tuning the parameter $\lambda$ and $\beta$}
        \label{fig:eval:tuning}
    \end{minipage}
    \hfill
    \begin{minipage}[b]{0.70\textwidth}
        \centering
        \resizebox{\textwidth}{!}{
        \begin{tabular}{ccccccc}
        \hline
        {Faithfulness metric} & {\AlgoName}  & {\AlgoName non active ($\beta$=0) } & {BRS+}  & {SP-Anchor} & {DT-Extract}  \\
        \hline
          acc         & 0.897 & 0.884 & 0.882 & 0.807 & 0.829 \\
          F1 score    & 0.741 & 0.690 & 0.709 & 0.205 & 0.357 \\
          recall      & 0.688 & 0.602 & 0.666 & 0.116 & 0.221 \\
          precision   & 0.803 & 0.809 & 0.757 & 0.877 & 0.924 \\
        \hline
        {Interpretability metric} &  &  &  &  &  \\
        \hline
          \# of rules                           & 24      & 29    & 57    & 24   &  24 \\
          average \# of conditions              & 4.375   & 2.413 & 3.859 & 7.875  & 5.375 \\
          max \# of conditions                  & 6       & 4     & 5     & 12    & 8 \\
        \end{tabular}
        }
        \captionof{table}{comparison of \AlgoName against baselines}
        \label{table:eval:overall}
    \end{minipage}
\end{minipage}


\Cref{table:eval:passive} shows the results of a comparison against more kinds of models other than decision sets by enlarging the dataset as preprocessing \citep{Bertini:2018:RuleMatrix} with the same number of new instances generated by \textbf{ADS}.
We observe that
\begin{itemize}
    \item \AlgoName is very competitive, while it only consists of a few human-readable rules. These rules are unordered and thus can be inspected separately, which makes \AlgoName (decision set method) superior for interpretability compared other models like decision list and tree.
    \AlgoName is comparable in terms of faithfulness with SVM and DNN which have massive parameters and are generally considered as not interpretable.
    \item Integrating the generation of synthetic instances into the algorithm gives a better improvement, while enlarging the dataset as preprocessing generally improves these passive methods only slightly. This also indicates the potential of designing an active model interpretation algorithm for SBRL, Random Forest, etc.
\end{itemize}



\begin{table}[h]
    \tiny
    \centering
    \resizebox{\textwidth}{!}{
    \begin{tabular}{ccccc ccccc}
    \hline
    {} & {\AlgoName} & {\textbf{ADS}+($\beta$=0) } & {\AlgoName ($\beta$=0) }   &  {BRS+/BRS}  & {SBRL+/SBRL}  & {CART+/CART} & {RF+/RF} & {SVM+/SVM} & {DNN+/DNN}  \\
    \hline

        acc                  & 0.897 & 0.895 & 0.884  & 0.882/ 0.868 & 0.903 / 0.901  & 0.911 / 0.910 & 0.932 / 0.929 & 0.915 / 0.913 & 0.943 / 0.935 \\
        F1 score             & 0.741 & 0.735 & 0.690  & 0.709 / 0.695 & 0.762 / 0.751 & 0.796 / 0.797 & 0.831 / 0.824 & 0.791 / 0.788 & 0.869 / 0.854 \\
        recall               & 0.688  & 0.678 & 0.602 &  0.666 / 0.703 & 0.720 / 0.699 & 0.807 / 0.820 & 0.778 / 0.773 & 0.747 / 0.755 & 0.872 / 0.875 \\
        precision            & 0.803 & 0.802 & 0.809  & 0.757 / 0.687 & 0.808 / 0.812 & 0.786 / 0.775 & 0.891 / 0.882 & 0.841 / 0.824 & 0.865 / 0.833 \\
        \# of rules
                            & 24  & 28 & 29 & 57 / 56  & 143 / 107 & 7741 / 4406 & 82850 / 44687  &
                            \begin{tabular}{@{}c@{}} 22689 / 10866  \\ \# of support vectors\end{tabular}
                            &
                            \begin{tabular}{@{}c@{}}  28310 (5 hidden layers)   \\ \# of weight parameters   \end{tabular}
                                                                                                \\
        average \# of conditions
                            & 4.375  & 2.928 &  2.413 & 3.859 /  3.663   &  1.853 /  1.682 & 8.664 / 8.869 &  9.322 / 9.571 & & \\
         maximum \# of conditions
                            & 6 & 4 & 4 &   5 / 5  & 2 / 2 & 12 / 12 & 12 / 12 & &\\
    \end{tabular}
    }
 \captionof{figure}{comparison of \AlgoName against more kinds of models}
 {%

 }
 \label{table:eval:passive}
\end{table}

\section{Conclusion}
\label{sec:conclusion}

In this paper, we present \textbf{ADS}, an active approach for model interpretation and demonstrate that integrating active-query into the algorithm is beneficial for finding better model interpretations, in terms of both faithfulness and interpretability. Though \AlgoName is only a simple algorithm and needs further polishing, our preliminary experimental results suggest that the active-query paradigm is promising for designing advanced model interpretation algorithms, especially iteration-based discrete optimization algorithms.

\paragraph{limitations}
The main limitations of \AlgoName is that its algorithm still appears to be too weak in terms of optimality. The greedy hill-climbing does not ensure convergence and taking the greedy action sometimes may not give the best result. It might make more sense to identify the best action to apply considering a longer range of iterations. Second, because of the generated synthetic instances, the estimate of the objective function $Q$ is biased by the local search procedure and the given dataset. A clever way to solve this issue is desired.
Last, the \AlgoName method, as well as most of the related methods we discussed, do not consider the role of uncertainty which is indeed very crucial for model interpretation.

\bibliographystyle{acmart}
{
\bibliography{paper,misc}


\begin{thebibliography}{00}


\ifx \showCODEN    \undefined \def \showCODEN     #1{\unskip}     \fi
\ifx \showDOI      \undefined \def \showDOI       #1{#1}\fi
\ifx \showISBNx    \undefined \def \showISBNx     #1{\unskip}     \fi
\ifx \showISBNxiii \undefined \def \showISBNxiii  #1{\unskip}     \fi
\ifx \showISSN     \undefined \def \showISSN      #1{\unskip}     \fi
\ifx \showLCCN     \undefined \def \showLCCN      #1{\unskip}     \fi
\ifx \shownote     \undefined \def \shownote      #1{#1}          \fi
\ifx \showarticletitle \undefined \def \showarticletitle #1{#1}   \fi
\ifx \showURL      \undefined \def \showURL       {\relax}        \fi
\providecommand\bibfield[2]{#2}
\providecommand\bibinfo[2]{#2}
\providecommand\natexlab[1]{#1}
\providecommand\showeprint[2][]{arXiv:#2}

\bibitem[\protect\citeauthoryear{Audibert and Bubeck}{Audibert and
  Bubeck}{2010}]%
        {audibert:2010:bmi_mab}
\bibfield{author}{\bibinfo{person}{Jean-Yves Audibert} {and}
  \bibinfo{person}{S{\'e}bastien Bubeck}.} \bibinfo{year}{2010}\natexlab{}.
\newblock \showarticletitle{Best arm identification in multi-armed bandits}.
\newblock


\bibitem[\protect\citeauthoryear{Bastani, Kim, and Bastani}{Bastani
  et~al\mbox{.}}{2017}]%
        {Bastani:2017:active_tree}
\bibfield{author}{\bibinfo{person}{Osbert Bastani}, \bibinfo{person}{Carolyn
  Kim}, {and} \bibinfo{person}{Hamsa Bastani}.}
  \bibinfo{year}{2017}\natexlab{}.
\newblock \showarticletitle{{Interpreting Blackbox Models via Model
  Extraction}}.
\newblock \bibinfo{journal}{{\em CoRR\/}}  \bibinfo{volume}{abs/1705.08504}
  (\bibinfo{year}{2017}).
\newblock
\showeprint[arxiv]{1705.08504}
\showURL{%
\url{http://arxiv.org/abs/1705.08504}}


\bibitem[\protect\citeauthoryear{Breiman and Shang}{Breiman and Shang}{1996}]%
        {Breiman:1996:born_again}
\bibfield{author}{\bibinfo{person}{Leo Breiman} {and} \bibinfo{person}{Nong
  Shang}.} \bibinfo{year}{1996}\natexlab{}.
\newblock \showarticletitle{Born again trees}.
\newblock  (\bibinfo{year}{1996}).
\newblock


\bibitem[\protect\citeauthoryear{Bryan, Nichol, Genovese, Schneider, Miller,
  and Wasserman}{Bryan et~al\mbox{.}}{2006}]%
        {NIPS:2005:straddle}
\bibfield{author}{\bibinfo{person}{Brent Bryan}, \bibinfo{person}{Robert~C.
  Nichol}, \bibinfo{person}{Christopher~R Genovese}, \bibinfo{person}{Jeff
  Schneider}, \bibinfo{person}{Christopher~J. Miller}, {and}
  \bibinfo{person}{Larry Wasserman}.} \bibinfo{year}{2006}\natexlab{}.
\newblock \showarticletitle{{Active Learning For Identifying Function Threshold
  Boundaries}}.
\newblock In \bibinfo{booktitle}{{\em Advances in Neural Information Processing
  Systems 18}}, \bibfield{editor}{\bibinfo{person}{Y.~Weiss},
  \bibinfo{person}{B.~Sch\"{o}lkopf}, {and} \bibinfo{person}{J.~C. Platt}}
  (Eds.). \bibinfo{publisher}{MIT Press}, \bibinfo{pages}{163--170}.
\newblock
\showURL{%
\url{http://papers.nips.cc/paper/2940-active-learning-for-identifying-function-threshold-boundaries.pdf}}


\bibitem[\protect\citeauthoryear{Craven}{Craven}{1996}]%
        {Craven:1996:TREPAN}
\bibfield{author}{\bibinfo{person}{Mark~William Craven}.}
  \bibinfo{year}{1996}\natexlab{}.
\newblock {\em \bibinfo{title}{{Extracting Comprehensible Models from Trained
  Neural Networks}}}.
\newblock \bibinfo{thesistype}{Ph.D. Dissertation}.
\newblock
\showISBNx{0-591-14495-6}
\newblock
\shownote{AAI9700774.}


\bibitem[\protect\citeauthoryear{Freitas}{Freitas}{2014}]%
        {Freitas_2014}
\bibfield{author}{\bibinfo{person}{Alex~A. Freitas}.}
  \bibinfo{year}{2014}\natexlab{}.
\newblock \showarticletitle{{Comprehensible classification models}}.
\newblock \bibinfo{journal}{{\em {ACM} {SIGKDD} Explorations Newsletter\/}}
  \bibinfo{volume}{15}, \bibinfo{number}{1} (\bibinfo{date}{mar}
  \bibinfo{year}{2014}), \bibinfo{pages}{1--10}.
\newblock
\showDOI{%
\url{https://doi.org/10.1145/2594473.2594475}}


\bibitem[\protect\citeauthoryear{Guidotti, Monreale, Ruggieri, Pedreschi,
  Turini, and Giannotti}{Guidotti et~al\mbox{.}}{2018}]%
        {LORE}
\bibfield{author}{\bibinfo{person}{Riccardo Guidotti}, \bibinfo{person}{Anna
  Monreale}, \bibinfo{person}{Salvatore Ruggieri}, \bibinfo{person}{Dino
  Pedreschi}, \bibinfo{person}{Franco Turini}, {and} \bibinfo{person}{Fosca
  Giannotti}.} \bibinfo{year}{2018}\natexlab{}.
\newblock \showarticletitle{{Local Rule-Based Explanations of Black Box
  Decision Systems}}.
\newblock \bibinfo{journal}{{\em CoRR\/}}  \bibinfo{volume}{abs/1805.10820}
  (\bibinfo{year}{2018}).
\newblock
\showeprint[arxiv]{1805.10820}
\showURL{%
\url{http://arxiv.org/abs/1805.10820}}


\bibitem[\protect\citeauthoryear{Huysmans, Dejaeger, Mues, Vanthienen, and
  Baesens}{Huysmans et~al\mbox{.}}{2011}]%
        {Huysmans_2011}
\bibfield{author}{\bibinfo{person}{Johan Huysmans}, \bibinfo{person}{Karel
  Dejaeger}, \bibinfo{person}{Christophe Mues}, \bibinfo{person}{Jan
  Vanthienen}, {and} \bibinfo{person}{Bart Baesens}.}
  \bibinfo{year}{2011}\natexlab{}.
\newblock \showarticletitle{{An empirical evaluation of the comprehensibility
  of decision table tree and rule based predictive models}}.
\newblock \bibinfo{journal}{{\em Decision Support Systems\/}}
  \bibinfo{volume}{51}, \bibinfo{number}{1} (\bibinfo{date}{apr}
  \bibinfo{year}{2011}), \bibinfo{pages}{141--154}.
\newblock
\showDOI{%
\url{https://doi.org/10.1016/j.dss.2010.12.003}}


\bibitem[\protect\citeauthoryear{Kalyanakrishnan, Tewari, Auer, and
  Stone}{Kalyanakrishnan et~al\mbox{.}}{2012}]%
        {kalyanakrishnan:2012:LUCB_PAC}
\bibfield{author}{\bibinfo{person}{Shivaram Kalyanakrishnan},
  \bibinfo{person}{Ambuj Tewari}, \bibinfo{person}{Peter Auer}, {and}
  \bibinfo{person}{Peter Stone}.} \bibinfo{year}{2012}\natexlab{}.
\newblock \showarticletitle{PAC Subset Selection in Stochastic Multi-armed
  Bandits.}. In \bibinfo{booktitle}{{\em ICML}}, Vol.~\bibinfo{volume}{12}.
  \bibinfo{pages}{655--662}.
\newblock


\bibitem[\protect\citeauthoryear{Kaufmann and Kalyanakrishnan}{Kaufmann and
  Kalyanakrishnan}{2013}]%
        {kaufmann:2013:information}
\bibfield{author}{\bibinfo{person}{Emilie Kaufmann} {and}
  \bibinfo{person}{Shivaram Kalyanakrishnan}.} \bibinfo{year}{2013}\natexlab{}.
\newblock \showarticletitle{Information complexity in bandit subset selection}.
  In \bibinfo{booktitle}{{\em Conference on Learning Theory}}.
  \bibinfo{pages}{228--251}.
\newblock


\bibitem[\protect\citeauthoryear{Kohavi}{Kohavi}{1996}]%
        {Kohavi:1996:Adult}
\bibfield{author}{\bibinfo{person}{Ron Kohavi}.}
  \bibinfo{year}{1996}\natexlab{}.
\newblock \showarticletitle{Scaling Up the Accuracy of Naive-Bayes Classifiers:
  A Decision-tree Hybrid}. In \bibinfo{booktitle}{{\em Proceedings of the
  Second International Conference on Knowledge Discovery and Data Mining}} {\em
  (\bibinfo{series}{KDD'96})}. \bibinfo{publisher}{AAAI Press},
  \bibinfo{pages}{202--207}.
\newblock
\showURL{%
\url{http://dl.acm.org/citation.cfm?id=3001460.3001502}}


\bibitem[\protect\citeauthoryear{Lakkaraju, Bach, and Leskovec}{Lakkaraju
  et~al\mbox{.}}{2016}]%
        {Lakkaraju:2016:IDS}
\bibfield{author}{\bibinfo{person}{Himabindu Lakkaraju},
  \bibinfo{person}{Stephen~H. Bach}, {and} \bibinfo{person}{Jure Leskovec}.}
  \bibinfo{year}{2016}\natexlab{}.
\newblock \showarticletitle{{Interpretable Decision Sets: A Joint Framework for
  Description and Prediction}}. In \bibinfo{booktitle}{{\em Proceedings of the
  22Nd ACM SIGKDD International Conference on Knowledge Discovery and Data
  Mining}} {\em (\bibinfo{series}{KDD '16})}. \bibinfo{publisher}{ACM},
  \bibinfo{address}{New York, NY, USA}, \bibinfo{pages}{1675--1684}.
\newblock
\showISBNx{978-1-4503-4232-2}
\showDOI{%
\url{https://doi.org/10.1145/2939672.2939874}}


\bibitem[\protect\citeauthoryear{Lakkaraju, Kamar, Caruana, and
  Leskovec}{Lakkaraju et~al\mbox{.}}{2017}]%
        {Jure:2017:BETA}
\bibfield{author}{\bibinfo{person}{Himabindu Lakkaraju}, \bibinfo{person}{Ece
  Kamar}, \bibinfo{person}{Rich Caruana}, {and} \bibinfo{person}{Jure
  Leskovec}.} \bibinfo{year}{2017}\natexlab{}.
\newblock \showarticletitle{{Interpretable {\&} Explorable Approximations of
  Black Box Models}}.
\newblock \bibinfo{journal}{{\em CoRR\/}}  \bibinfo{volume}{abs/1707.01154}
  (\bibinfo{year}{2017}).
\newblock
\showeprint[arxiv]{1707.01154}
\showURL{%
\url{http://arxiv.org/abs/1707.01154}}


\bibitem[\protect\citeauthoryear{Lakkaraju, Kamar, Caruana, and
  Leskovec}{Lakkaraju et~al\mbox{.}}{2019}]%
        {lakkaraju:2019:faithful}
\bibfield{author}{\bibinfo{person}{Himabindu Lakkaraju}, \bibinfo{person}{Ece
  Kamar}, \bibinfo{person}{Rich Caruana}, {and} \bibinfo{person}{Jure
  Leskovec}.} \bibinfo{year}{2019}\natexlab{}.
\newblock \showarticletitle{{Faithful and Customizable Explanations of Black
  Box Models}}.
\newblock \bibinfo{journal}{{\em AIES\/}} (\bibinfo{year}{2019}).
\newblock


\bibitem[\protect\citeauthoryear{MacKay}{MacKay}{1992}]%
        {MacKay_1992}
\bibfield{author}{\bibinfo{person}{David J.~C. MacKay}.}
  \bibinfo{year}{1992}\natexlab{}.
\newblock \showarticletitle{{Information-Based Objective Functions for Active
  Data Selection}}.
\newblock \bibinfo{journal}{{\em Neural Computation\/}} \bibinfo{volume}{4},
  \bibinfo{number}{4} (\bibinfo{date}{jul} \bibinfo{year}{1992}),
  \bibinfo{pages}{590--604}.
\newblock
\showDOI{%
\url{https://doi.org/10.1162/neco.1992.4.4.590}}


\bibitem[\protect\citeauthoryear{Ming, Qu, and Bertini}{Ming
  et~al\mbox{.}}{2018}]%
        {Bertini:2018:RuleMatrix}
\bibfield{author}{\bibinfo{person}{Yao Ming}, \bibinfo{person}{Huamin Qu},
  {and} \bibinfo{person}{Enrico Bertini}.} \bibinfo{year}{2018}\natexlab{}.
\newblock \showarticletitle{{RuleMatrix: Visualizing and Understanding
  Classifiers with Rules}}.
\newblock \bibinfo{journal}{{\em CoRR\/}}  \bibinfo{volume}{abs/1807.06228}
  (\bibinfo{year}{2018}).
\newblock
\showeprint[arxiv]{1807.06228}
\showURL{%
\url{http://arxiv.org/abs/1807.06228}}


\bibitem[\protect\citeauthoryear{Pa{\c{c}}aci, Johnson, McKeever, and
  Hamfelt}{Pa{\c{c}}aci et~al\mbox{.}}{2019}]%
        {Hamfelt:2019:RICE}
\bibfield{author}{\bibinfo{person}{G{\"{o}}rkem Pa{\c{c}}aci},
  \bibinfo{person}{David Johnson}, \bibinfo{person}{Steve McKeever}, {and}
  \bibinfo{person}{Andreas Hamfelt}.} \bibinfo{year}{2019}\natexlab{}.
\newblock \showarticletitle{{Why did you do that?: Explaining black box models
  with Inductive Synthesis}}.
\newblock \bibinfo{journal}{{\em CoRR\/}}  \bibinfo{volume}{abs/1904.09273}
  (\bibinfo{year}{2019}).
\newblock
\showeprint[arxiv]{1904.09273}
\showURL{%
\url{http://arxiv.org/abs/1904.09273}}


\bibitem[\protect\citeauthoryear{Pedreschi, Giannotti, Guidotti, Monreale,
  Pappalardo, Ruggieri, and Turini}{Pedreschi et~al\mbox{.}}{2018}]%
        {Dino:2018:local_first}
\bibfield{author}{\bibinfo{person}{Dino Pedreschi}, \bibinfo{person}{Fosca
  Giannotti}, \bibinfo{person}{Riccardo Guidotti}, \bibinfo{person}{Anna
  Monreale}, \bibinfo{person}{Luca Pappalardo}, \bibinfo{person}{Salvatore
  Ruggieri}, {and} \bibinfo{person}{Franco Turini}.}
  \bibinfo{year}{2018}\natexlab{}.
\newblock \showarticletitle{{Open the Black Box Data-Driven Explanation of
  Black Box Decision Systems}}.
\newblock \bibinfo{journal}{{\em CoRR\/}}  \bibinfo{volume}{abs/1806.09936}
  (\bibinfo{year}{2018}).
\newblock
\showeprint[arxiv]{1806.09936}
\showURL{%
\url{http://arxiv.org/abs/1806.09936}}


\bibitem[\protect\citeauthoryear{Ribeiro, Singh, and Guestrin}{Ribeiro
  et~al\mbox{.}}{2016}]%
        {Ribeiro_2016}
\bibfield{author}{\bibinfo{person}{Marco~Tulio Ribeiro},
  \bibinfo{person}{Sameer Singh}, {and} \bibinfo{person}{Carlos Guestrin}.}
  \bibinfo{year}{2016}\natexlab{}.
\newblock \showarticletitle{{Why Should I Trust You?}}. In
  \bibinfo{booktitle}{{\em Proceedings of the 22nd {ACM} {SIGKDD} International
  Conference on Knowledge Discovery and Data Mining - {KDD} 16}}.
  \bibinfo{publisher}{{ACM} Press}.
\newblock
\showDOI{%
\url{https://doi.org/10.1145/2939672.2939778}}


\bibitem[\protect\citeauthoryear{Ribeiro, Singh, and Guestrin}{Ribeiro
  et~al\mbox{.}}{2018}]%
        {Ribeiro:2018:Anchor}
\bibfield{author}{\bibinfo{person}{Marco~T{\'u}lio Ribeiro},
  \bibinfo{person}{Sameer Singh}, {and} \bibinfo{person}{Carlos Guestrin}.}
  \bibinfo{year}{2018}\natexlab{}.
\newblock \showarticletitle{{Anchors: High-Precision Model-Agnostic
  Explanations}}. In \bibinfo{booktitle}{{\em AAAI}}.
\newblock


\bibitem[\protect\citeauthoryear{Sushil, Suster, and Daelemans}{Sushil
  et~al\mbox{.}}{2018}]%
        {Sushil:2018:RuleInduction}
\bibfield{author}{\bibinfo{person}{Madhumita Sushil}, \bibinfo{person}{Simon
  Suster}, {and} \bibinfo{person}{Walter Daelemans}.}
  \bibinfo{year}{2018}\natexlab{}.
\newblock \showarticletitle{{Rule induction for global explanation of trained
  models}}.
\newblock \bibinfo{journal}{{\em CoRR\/}}  \bibinfo{volume}{abs/1808.09744}
  (\bibinfo{year}{2018}).
\newblock
\showeprint[arxiv]{1808.09744}
\showURL{%
\url{http://arxiv.org/abs/1808.09744}}


\bibitem[\protect\citeauthoryear{Wang, Rudin, Doshi-Velez, Liu, Klampfl, and
  MacNeille}{Wang et~al\mbox{.}}{2017}]%
        {Wang:2017:BRS}
\bibfield{author}{\bibinfo{person}{Tong Wang}, \bibinfo{person}{Cynthia Rudin},
  \bibinfo{person}{Finale Doshi-Velez}, \bibinfo{person}{Yimin Liu},
  \bibinfo{person}{Erica Klampfl}, {and} \bibinfo{person}{Perry MacNeille}.}
  \bibinfo{year}{2017}\natexlab{}.
\newblock \showarticletitle{{A Bayesian Framework for Learning Rule Sets for
  Interpretable Classification}}.
\newblock \bibinfo{journal}{{\em Journal of Machine Learning Research\/}}
  \bibinfo{volume}{18}, \bibinfo{number}{70} (\bibinfo{year}{2017}),
  \bibinfo{pages}{1--37}.
\newblock
\showURL{%
\url{http://jmlr.org/papers/v18/16-003.html}}


\bibitem[\protect\citeauthoryear{Yang, Rudin, and Seltzer}{Yang
  et~al\mbox{.}}{2017}]%
        {Yang:2017:SBRL}
\bibfield{author}{\bibinfo{person}{Hongyu Yang}, \bibinfo{person}{Cynthia
  Rudin}, {and} \bibinfo{person}{Margo Seltzer}.}
  \bibinfo{year}{2017}\natexlab{}.
\newblock \showarticletitle{{Scalable Bayesian Rule Lists}}. In
  \bibinfo{booktitle}{{\em Proceedings of the 34th International Conference on
  Machine Learning - Volume 70}} {\em (\bibinfo{series}{ICML'17})}.
  \bibinfo{publisher}{JMLR.org}, \bibinfo{pages}{3921--3930}.
\newblock
\showURL{%
\url{http://dl.acm.org/citation.cfm?id=3305890.3306086}}


\end{thebibliography}



\begin{thebibliography}{00}


\ifx \showCODEN    \undefined \def \showCODEN     #1{\unskip}     \fi
\ifx \showDOI      \undefined \def \showDOI       #1{#1}\fi
\ifx \showISBNx    \undefined \def \showISBNx     #1{\unskip}     \fi
\ifx \showISBNxiii \undefined \def \showISBNxiii  #1{\unskip}     \fi
\ifx \showISSN     \undefined \def \showISSN      #1{\unskip}     \fi
\ifx \showLCCN     \undefined \def \showLCCN      #1{\unskip}     \fi
\ifx \shownote     \undefined \def \shownote      #1{#1}          \fi
\ifx \showarticletitle \undefined \def \showarticletitle #1{#1}   \fi
\ifx \showURL      \undefined \def \showURL       {\relax}        \fi
\providecommand\bibfield[2]{#2}
\providecommand\bibinfo[2]{#2}
\providecommand\natexlab[1]{#1}
\providecommand\showeprint[2][]{arXiv:#2}

\bibitem[\protect\citeauthoryear{Angelino, Larus-Stone, Alabi, Seltzer, and
  Rudin}{Angelino et~al\mbox{.}}{2017}]%
        {Angelino:2017:CORELS}
\bibfield{author}{\bibinfo{person}{Elaine Angelino}, \bibinfo{person}{Nicholas
  Larus-Stone}, \bibinfo{person}{Daniel Alabi}, \bibinfo{person}{Margo
  Seltzer}, {and} \bibinfo{person}{Cynthia Rudin}.}
  \bibinfo{year}{2017}\natexlab{}.
\newblock \showarticletitle{{Learning Certifiably Optimal Rule Lists}}. In
  \bibinfo{booktitle}{{\em Proceedings of the 23rd ACM SIGKDD International
  Conference on Knowledge Discovery and Data Mining}} {\em
  (\bibinfo{series}{KDD '17})}. \bibinfo{publisher}{ACM}, \bibinfo{address}{New
  York, NY, USA}, \bibinfo{pages}{35--44}.
\newblock
\showISBNx{978-1-4503-4887-4}
\showDOI{%
\url{https://doi.org/10.1145/3097983.3098047}}


\bibitem[\protect\citeauthoryear{Bastani, Kim, and Bastani}{Bastani
  et~al\mbox{.}}{2017}]%
        {Bastani:2017:active_tree}
\bibfield{author}{\bibinfo{person}{Osbert Bastani}, \bibinfo{person}{Carolyn
  Kim}, {and} \bibinfo{person}{Hamsa Bastani}.}
  \bibinfo{year}{2017}\natexlab{}.
\newblock \showarticletitle{{Interpreting Blackbox Models via Model
  Extraction}}.
\newblock \bibinfo{journal}{{\em CoRR\/}}  \bibinfo{volume}{abs/1705.08504}
  (\bibinfo{year}{2017}).
\newblock
\showeprint[arxiv]{1705.08504}
\showURL{%
\url{http://arxiv.org/abs/1705.08504}}


\bibitem[\protect\citeauthoryear{Craven}{Craven}{1996}]%
        {Craven:1996:TREPAN}
\bibfield{author}{\bibinfo{person}{Mark~William Craven}.}
  \bibinfo{year}{1996}\natexlab{}.
\newblock {\em \bibinfo{title}{{Extracting Comprehensible Models from Trained
  Neural Networks}}}.
\newblock \bibinfo{thesistype}{Ph.D. Dissertation}.
\newblock
\showISBNx{0-591-14495-6}
\newblock
\shownote{AAI9700774.}


\bibitem[\protect\citeauthoryear{Freitas}{Freitas}{2014}]%
        {Freitas_2014}
\bibfield{author}{\bibinfo{person}{Alex~A. Freitas}.}
  \bibinfo{year}{2014}\natexlab{}.
\newblock \showarticletitle{{Comprehensible classification models}}.
\newblock \bibinfo{journal}{{\em {ACM} {SIGKDD} Explorations Newsletter\/}}
  \bibinfo{volume}{15}, \bibinfo{number}{1} (\bibinfo{date}{mar}
  \bibinfo{year}{2014}), \bibinfo{pages}{1--10}.
\newblock
\showDOI{%
\url{https://doi.org/10.1145/2594473.2594475}}


\bibitem[\protect\citeauthoryear{Huysmans, Dejaeger, Mues, Vanthienen, and
  Baesens}{Huysmans et~al\mbox{.}}{2011}]%
        {Huysmans_2011}
\bibfield{author}{\bibinfo{person}{Johan Huysmans}, \bibinfo{person}{Karel
  Dejaeger}, \bibinfo{person}{Christophe Mues}, \bibinfo{person}{Jan
  Vanthienen}, {and} \bibinfo{person}{Bart Baesens}.}
  \bibinfo{year}{2011}\natexlab{}.
\newblock \showarticletitle{{An empirical evaluation of the comprehensibility
  of decision table tree and rule based predictive models}}.
\newblock \bibinfo{journal}{{\em Decision Support Systems\/}}
  \bibinfo{volume}{51}, \bibinfo{number}{1} (\bibinfo{date}{apr}
  \bibinfo{year}{2011}), \bibinfo{pages}{141--154}.
\newblock
\showDOI{%
\url{https://doi.org/10.1016/j.dss.2010.12.003}}


\bibitem[\protect\citeauthoryear{Lakkaraju, Bach, and Leskovec}{Lakkaraju
  et~al\mbox{.}}{2016}]%
        {Lakkaraju:2016:IDS}
\bibfield{author}{\bibinfo{person}{Himabindu Lakkaraju},
  \bibinfo{person}{Stephen~H. Bach}, {and} \bibinfo{person}{Jure Leskovec}.}
  \bibinfo{year}{2016}\natexlab{}.
\newblock \showarticletitle{{Interpretable Decision Sets: A Joint Framework for
  Description and Prediction}}. In \bibinfo{booktitle}{{\em Proceedings of the
  22Nd ACM SIGKDD International Conference on Knowledge Discovery and Data
  Mining}} {\em (\bibinfo{series}{KDD '16})}. \bibinfo{publisher}{ACM},
  \bibinfo{address}{New York, NY, USA}, \bibinfo{pages}{1675--1684}.
\newblock
\showISBNx{978-1-4503-4232-2}
\showDOI{%
\url{https://doi.org/10.1145/2939672.2939874}}


\bibitem[\protect\citeauthoryear{Lakkaraju, Kamar, Caruana, and
  Leskovec}{Lakkaraju et~al\mbox{.}}{2019}]%
        {lakkaraju:2019:faithful}
\bibfield{author}{\bibinfo{person}{Himabindu Lakkaraju}, \bibinfo{person}{Ece
  Kamar}, \bibinfo{person}{Rich Caruana}, {and} \bibinfo{person}{Jure
  Leskovec}.} \bibinfo{year}{2019}\natexlab{}.
\newblock \showarticletitle{{Faithful and Customizable Explanations of Black
  Box Models}}.
\newblock \bibinfo{journal}{{\em AIES\/}} (\bibinfo{year}{2019}).
\newblock


\bibitem[\protect\citeauthoryear{Ming, Qu, and Bertini}{Ming
  et~al\mbox{.}}{2018}]%
        {Bertini:2018:RuleMatrix}
\bibfield{author}{\bibinfo{person}{Yao Ming}, \bibinfo{person}{Huamin Qu},
  {and} \bibinfo{person}{Enrico Bertini}.} \bibinfo{year}{2018}\natexlab{}.
\newblock \showarticletitle{{RuleMatrix: Visualizing and Understanding
  Classifiers with Rules}}.
\newblock \bibinfo{journal}{{\em CoRR\/}}  \bibinfo{volume}{abs/1807.06228}
  (\bibinfo{year}{2018}).
\newblock
\showeprint[arxiv]{1807.06228}
\showURL{%
\url{http://arxiv.org/abs/1807.06228}}


\bibitem[\protect\citeauthoryear{Pedreschi, Giannotti, Guidotti, Monreale,
  Pappalardo, Ruggieri, and Turini}{Pedreschi et~al\mbox{.}}{2018}]%
        {Dino:2018:local_first}
\bibfield{author}{\bibinfo{person}{Dino Pedreschi}, \bibinfo{person}{Fosca
  Giannotti}, \bibinfo{person}{Riccardo Guidotti}, \bibinfo{person}{Anna
  Monreale}, \bibinfo{person}{Luca Pappalardo}, \bibinfo{person}{Salvatore
  Ruggieri}, {and} \bibinfo{person}{Franco Turini}.}
  \bibinfo{year}{2018}\natexlab{}.
\newblock \showarticletitle{{Open the Black Box Data-Driven Explanation of
  Black Box Decision Systems}}.
\newblock \bibinfo{journal}{{\em CoRR\/}}  \bibinfo{volume}{abs/1806.09936}
  (\bibinfo{year}{2018}).
\newblock
\showeprint[arxiv]{1806.09936}
\showURL{%
\url{http://arxiv.org/abs/1806.09936}}


\bibitem[\protect\citeauthoryear{Stackexchange}{Stackexchange}{2011}]%
        {StackExchange:2011:HLI}
\bibfield{author}{\bibinfo{person}{Stackexchange}.}
  \bibinfo{year}{2011}\natexlab{}.
\newblock \bibinfo{title}{How to link images relatively in Inkscape?}
\newblock   (\bibinfo{year}{2011}).
\newblock
\newblock
\shownote{\url{http://graphicdesign.stackexchange.com/questions/4906/how-to-link-images-relatively-in-inkscape}.}


\bibitem[\protect\citeauthoryear{Sun, Wei, and Kaufman}{Sun
  et~al\mbox{.}}{2016}]%
        {Sun:2016:MVP}
\bibfield{author}{\bibinfo{person}{Qi Sun}, \bibinfo{person}{Li-Yi Wei}, {and}
  \bibinfo{person}{Arie Kaufman}.} \bibinfo{year}{2016}\natexlab{}.
\newblock \showarticletitle{Mapping Virtual and Physical Reality}.
\newblock \bibinfo{journal}{{\em ACM Trans. Graph.\/}} \bibinfo{volume}{35},
  \bibinfo{number}{4}, Article \bibinfo{articleno}{64} (\bibinfo{date}{July}
  \bibinfo{year}{2016}), \bibinfo{numpages}{12}~pages.
\newblock
\showISSN{0730-0301}
\showDOI{%
\url{https://doi.org/10.1145/2897824.2925883}}


\bibitem[\protect\citeauthoryear{Wang, Rudin, Doshi-Velez, Liu, Klampfl, and
  MacNeille}{Wang et~al\mbox{.}}{2017}]%
        {Wang:2017:BRS}
\bibfield{author}{\bibinfo{person}{Tong Wang}, \bibinfo{person}{Cynthia Rudin},
  \bibinfo{person}{Finale Doshi-Velez}, \bibinfo{person}{Yimin Liu},
  \bibinfo{person}{Erica Klampfl}, {and} \bibinfo{person}{Perry MacNeille}.}
  \bibinfo{year}{2017}\natexlab{}.
\newblock \showarticletitle{{A Bayesian Framework for Learning Rule Sets for
  Interpretable Classification}}.
\newblock \bibinfo{journal}{{\em Journal of Machine Learning Research\/}}
  \bibinfo{volume}{18}, \bibinfo{number}{70} (\bibinfo{year}{2017}),
  \bibinfo{pages}{1--37}.
\newblock
\showURL{%
\url{http://jmlr.org/papers/v18/16-003.html}}


\bibitem[\protect\citeauthoryear{Yang, Rudin, and Seltzer}{Yang
  et~al\mbox{.}}{2017}]%
        {Yang:2017:SBRL}
\bibfield{author}{\bibinfo{person}{Hongyu Yang}, \bibinfo{person}{Cynthia
  Rudin}, {and} \bibinfo{person}{Margo Seltzer}.}
  \bibinfo{year}{2017}\natexlab{}.
\newblock \showarticletitle{{Scalable Bayesian Rule Lists}}. In
  \bibinfo{booktitle}{{\em Proceedings of the 34th International Conference on
  Machine Learning - Volume 70}} {\em (\bibinfo{series}{ICML'17})}.
  \bibinfo{publisher}{JMLR.org}, \bibinfo{pages}{3921--3930}.
\newblock
\showURL{%
\url{http://dl.acm.org/citation.cfm?id=3305890.3306086}}


\end{thebibliography}
}

\appendix
\normalsize

\ifthenelse{\equal{\final}{0}}
{
\clearpage
\pagenumbering{roman}

\section{Appendix}

\begin{description}
\item[More on related work]
{
    For example, \cite{Bertini:2018:RuleMatrix} first estimate $P(X)$ and then sample an arbitray large dataset. \cite{Hamfelt:2019:RICE} exhaustively enumerates the input space and then pick up the most "important" instances. \cite{Sushil:2018:RuleInduction} takes an extra step to preprocess the dataset by extracting important features. All these approaches are followed by a learning algorithm which is no different from supervised rule induction.
    We want to emphasize that the approaches do utilize the active-query property, but in a separate procedure away from the main learning algorithm and thus can lead to suboptimal results.

}
\item[An a visualizab 2-d synthetic dataset]
{

We first apply our method to a simple 2-d synthetic dataset which is adapted from \cite{NIPS:2005:straddle} in the context of testing active sampling technique.

We find that, our \AlgoName will sample more new instances near the decision boundary. This is consistent with our orignal goal that since the true decision boundary might not be revealed in the given dataset, we wish to actively generate new data instances that will reveal this decision boundary.

\begin{figure}[htb]
  \centering
  \subfloat[IDS -> will change to MUSE]{
    \label{fig:synthetic:IDS}
    \includegraphics[width=0.2\linewidth]{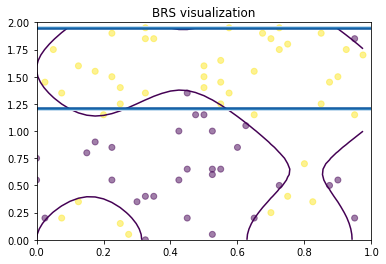}
  }
  \subfloat[our method: non active -> will change to Dino's method]{
    \label{fig:synthetic:non_active}
    \includegraphics[width=0.2\linewidth]{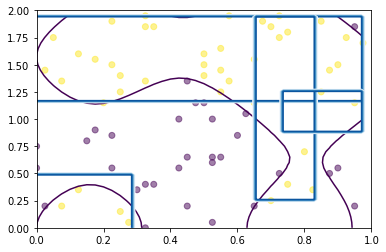}
  }
  \subfloat[DT-Extract]{
    \label{fig:synthetic:non_active}
    \includegraphics[width=0.2\linewidth]{figs/raster/synthetic/DT-Extract.png}
  }
  \subfloat[our method: active]{
    \label{fig:synthetic:our_active}
    \includegraphics[width=0.2\linewidth]{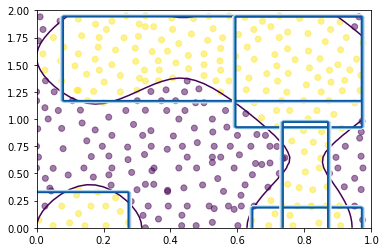}
  }
 \Caption{Visualzation on the toy synthetic dataset}
 {%
 We compare our method, with three representative methods from each category: (a) passive methods: we use Interpretable decision set (Will change to MUSE) (b) the active, bottom-up method will change to Dino's clustering method (c) the active, top-down method: DT-Extract, build a decision tree and generate a fixed number of new instances to query for each node split. (d). Our method.
 }
 \label{fig:synthetic:overall}
\end{figure}

\begin{table}[htb!]
  \begin{center}
    \label{tab:synthetic:overall}
    \begin{tabular}{cccccc}
    \hline
    {Faifulness metric} & {IDS} & Local-first & DT-Extract & {ours non active} & {ours active}  \\
    \hline
        acc  & 0.72 &  & 0.933 & 0.800 & 0.946\\
        f1 score  & 0.7586 &  & 0.935  & 0.814 & 0.945 \\
        recall  & 0.8148 &  & 0.972  & 0.891 & 0.945 \\
        precision  & 0.7096 &  & 0.9 & 0.75 & 0.945 \\
    \hline
    {Interpretability metric} & {IDS} & Local-first & DT-Extract  & {ours non active} & {ours active} \\
    \hline
        number of rules  & 4 & & 8 & 4 & 5\\
        Other  &   &  &  \\
    \end{tabular}
    \caption{comparison of methods in this toy synthetic dataset}
    {
    We compare our algorithm \AlgoName with the baselines on various metrics of faithfulness and interpretability on the synthetic 2-d toy dataset.
    }
  \end{center}
\end{table}
}
\item[Experimental setting]
{
    For some methods that require discretizing of the features, we bin numerical attributes using MDL criteria for these methods.
}
\item[Comparing Generation Scheme]
{
}

\end{description}

}
{}

\end{document}